\documentclass[journal]{IEEEtran}

\usepackage{subfigure}
\usepackage{url}
\usepackage{amsmath}
\usepackage{mathptmx}
\usepackage{mathtools}
\usepackage{multirow}
\usepackage{multicol}
\usepackage{changepage}
\usepackage{algorithm}
\usepackage{algorithmic}
\usepackage{balance}
\usepackage{comment}
\usepackage{tabularx}
\usepackage{tabulary}
\usepackage{graphicx}
\usepackage{caption}
\usepackage{url}
\usepackage[export]{adjustbox}

\title{Efficient and accurate object detection with simultaneous classification and tracking}
\author{Xuesong Li, Jose Guivant, ~\IEEEmembership{Member,~IEEE,}\\
\thanks{X. Li (e-mail: xuesong.li@unsw.edu.au); J. Guivant (e-mail: j.guivant@unsw.edu.au). All above authors are with the School of Mechanical Engineering, University of New South Wales, Sydney, NSW 2052, AU}
}
\date{1/11/2018}

\begin{document}
\maketitle

\begin{abstract}

Interacting with the environment, such as object detection and tracking, is a crucial ability of mobile robots. Besides high accuracy, efficiency in terms of processing effort and energy consumption are also desirable. To satisfy both requirements, we propose a detection framework based on simultaneous classification and tracking in the point stream. In this framework, a tracker performs data association in sequences of the point cloud, guiding the detector to avoid redundant processing (i.e. classifying already-known objects). For objects whose classification is not sufficiently certain, a fusion model is designed to fuse selected key observations that provide different perspectives across the tracking span. Therefore, performance (accuracy and efficiency of detection) can be enhanced. This method is particularly suitable for detecting and tracking moving objects, a process which would require expensive computations if solved using conventional procedures. Experiments were conducted on the benchmark dataset, and the results showed that the proposed method outperforms original tracking-by-detection approaches in both efficiency and accuracy.

\end{abstract}
\begin{IEEEkeywords}
Moving object tracking, object detection, data association
\end{IEEEkeywords}

\IEEEpeerreviewmaketitle

\section{Introduction}\label{introduction}

\IEEEPARstart{M}{ulti-object} detection and tracking (MODT) are the critical pieces of information for autonomous robots in diverse applications, such as driving \cite{urmson2008autonomous} and home assistance \cite{ozkil2009service}, as the stationary or moving objects are key components with which robots interact in the scene. Multi-object detection (MOD) searches for objects of interest in a single frame, including their location and instance category, while multi-object tracking (MOT) links these objects across consecutive temporal frames and infers their motion behaviour. Currently, tracking-by-detection is the main paradigm for MODT \cite{SORT, baseline_tracking, Osep17ICRA, mono_3d_pmpf, doi:10.1177/0278364915593399, Lee2018RealTimeOT}. The detector is fully launched to process every frame of image or point cloud, and the results are passed to the tracker for data association and state estimation. (In this work one point cloud represents a 3D scan of the environment around the robot.) The tracker depends heavily on the detector, while the detector benefits marginally from the tracker. This kind of mechanism can involve a large number of redundant processing. The detector is repeatedly invoked to recognize objects which have already been tracked. In addition, valuable information, such as the confidence probability distribution about object class, is usually discarded by manually setting a winner-take-all threshold.

When humans perceive our surroundings, we do not repeatedly identify the same objects. Instead, we spend substantial energy on recognizing new objects when they first appear. Then we use a small amount of effort to track them. Therefore, the workload of the brain is significantly decreased, and its resources can be used for other purposes. For example, after identifying all the participants in a traffic scenario, the driver of an automobile would effortlessly track them and pay more attention to new objects or abnormal situations. The driver does not need to be in a very alert state to re-detect every detail in the scenario. Based on this observation, we argue that MOD and MOT should be mutually beneficial and can be assembled to perform MODT better than a conventional tracking-by-detection pipeline. To realise this idea, we propose a new MODT framework that elegantly integrates the detector and tracker to make detection more efficient and accurate.

In the proposed MODT framework, a two-stage detector \cite{my_efficient_paper}, which consists of point cloud segmentation and proposal classification, is used to detect objects in the form of 3D bounding boxes, and an extended Kalman filter (EKF) \cite{kalman_filter} is used to track every detected object. Point cloud segmentation efficiently identifies objects of interest, here called proposals, and it provides accurate geometrical location about 3D proposals. The proposals are further separated by a classification model into different categories, such as background objects, pedestrians, and cars. The proposal classification model is based on a deep neural network, and it requires much more computation cost than the point segmentation method. The function with low computation cost, the segmentation module, is launched for every 3D frame of the point cloud for securing the detection of new objects and for tracking them. The expensive function, classification module, is triggered only for objects that have not yet been classified, and the tracker would propagate the class or identity of recognised objects to subsequent frames. Therefore, the time dedicated to running the classification process is reduced significantly and the detection function becomes much more efficient than in the conventional MODT mechanism. In our MODT framework, a finite-state machine (FSM) is used to model how the state of every proposal ,or object, changes.

For objects about which we are not certain, we humans try to collect more information about them either by passively waiting for them to expose different views or by actively moving to change our angle of observation. This process continues until we identify the object with confidence. To simulate the process of improving the accuracy of detection, a fusion model is proposed to connect the classification probability of some keyframes across the temporal domain. These keyframes are selected if they meet our criteria for statistical independence. That is, the keyframes represent highly different views of the same object and carry independent information. Using the fusion model increases the classification cost slightly, but the total cost is much lower than conducting full detection.

Our aim is not to invent a new tracker and detector but to explore how an integrated tracker will improve detection in terms of both accuracy and efficiency. There are two main contributions of our work. The first is our proposed MODT framework, which elegantly combines the MOD and MOT to perform the MODT function at a lower computation cost and with higher detection accuracy. The second contribution is that we conduct a substantial number of experiments with the KITTI tracking dataset \cite{geiger2012we} to validate how the quality of the tracker and detector can affect the efficiency and accuracy of the MODT framework. The experiments are designed with different combinations of ideal and real tracker and detector.

The rest of the paper is organized as follows. Section \ref{literature_review} introduces the related work, followed by Section \ref{detector_trackor} which illustrates the adopted detector and tracker. The efficient and accurate MODT frameworks are explained in the Sections \ref{efficient_detection}  and \ref{accurate_detection}, respectively. Experiments using the proposed method are described in Section \ref{experiments_part}. Section \ref{conclusion_part} concludes our work and summarizes the findings.

\section{Literature Review}\label{literature_review}

Three-dimensional object detection is a very important part of MODT, and its structure and output also affect the mechanism of MODT. The literature on 3D object detection is reviewed first, followed by consideration of related works about multi-object tracking.

\subsection{Three-dimensional object detection}
The conventional pipeline of 3D object detection comprises two main steps: point cloud segmentation and segment classification \cite{Himmelsbach_2010, graph_ground,pfrsegmen, fast_segmentation,5979818, 5979636, wang2015voting}. Point cloud segmentation is usually based on some assumptions that the ground can be approximately fitted with one plane, and 3D objects are sufficiently separated. Douillard et al. \cite{5979818} proposed a Gaussian process of incremental sample consensus to reject ground points, and the remaining non-ground points were segmented into different clusters. As for segment classification, Teichman et al. \cite{5979636} used a segment classified and a holistic classifier to classify feature vectors. The results were combined in boosting the framework with a discrete Bayes filter. Conventional detection methods are based on rules instead of being data-driven, so they usually require a small annotated dataset. Moreover, they can be very efficient and are applicable for small robots with limited processing capabilities.

With the prevalence of convolutional neural network (CNN) in object detection, several related 3D object detection approaches were proposed \cite{mutliviews, avod_3d, DBLP:journals/corr/abs-1711-06396, myownpaper, s18103337}. 2D CNN has achieved tremendous success in computer vision tasks, and it can be implemented efficiently. Because of this, some 3D object detection methods \cite{mutliviews, avod_3d} transform 3D data into 2D images and employ 2D CNN directly for detection. Other methods \cite{DBLP:journals/corr/abs-1711-06396, myownpaper, s18103337} take 3D data as input without transforming it by using PointNet \cite{DBLP:journals/corr/QiSMG16, qi2017pointnetplusplus} or 3D sparse CNN \cite{SubmanifoldSparseConvNet, DBLP:journals/corr/Graham15}. Zhou et al. \cite{DBLP:journals/corr/abs-1711-06396} used PointNet to extract features from the low-resolution raw point cloud for each voxel, then used a shallow 3D CNN network to convert 3D feature maps into 2D feature maps for 3D object detection. Li. et al. \cite{myownpaper} designed multiple layers of sparse CNN to learn 3D feature maps from sparse 3D data to detect 3D object directly. These methods are based on deep neural networks, and they usually require an intensive computation unit, such as graphic processing unit (GPU).

The detector used in our framework follows a conventional detection structure, and it includes point cloud segmentation and segment classification. This makes the detector efficient and applicable to robots that have limited computation resources, as most small robots cannot support GPU to use deep learning models. Instead of using a simple machine model for classification, our detector uses PointNet \cite{DBLP:journals/corr/QiSMG16}, a deep learning model, to perform the classification task to improve detection accuracy. However, PointNet is launched at a low frequency to guarantee efficiency.

\subsection{Multi-object tracking}
Most 3D MOT follows the tracking-by-detection paradigm, where the detector is applied fully to every frame and finds objects in the form of a bounding box. Then the tracker is used to manage the tasks of estimating the state, associating data, starting a new tracklet, and eliminating an existing tracklet \cite{SORT, baseline_tracking, Osep17ICRA, mono_3d_pmpf, doi:10.1177/0278364915593399, Lee2018RealTimeOT}. State estimation is usually done with Bayesian filtering methods, such as particle filters \cite{Moral1997NonlinearFI}, EKF \cite{kalman_filter}, and Gaussian processes \cite{1640765}. Issues of data association can be addressed by using a bipartite graph matching algorithm, such as the Hungarian algorithm \cite{Kuhn55thehungarian}. Osep et al. \cite{Osep17ICRA} used a conditional random field (CRF) to fuse image-based detection results and 3D proposals to select a set of desirable detection observations. Then, objects in both the image domain and real-world space were tracked using an EKF in a joint 2D-3D state vector. The size difference between detected and tracked objects and the intersection over union (IoU) of 3D bounding boxes were calculated with metrics similar to data association. Scheidegger et al. \cite{mono_3d_pmpf} used CNN to detect objects in the image plane and their corresponding distance to the camera. Then they used the Poisson multi-Bernoulli mixture filter \cite{Poisson_multi_Bernoulli} to track the objects in 3D space. Weng et al. \cite{baseline_tracking} used PointRCNN \cite{Shi_2019_CVPR} to generate 3D object bounding boxes, and they used the 3D EKF and IoU-based Hungarian algorithms to tracking the 3D objects.

The distance-based bipartite graph matching algorithm is usually very efficient. However, in some complicated scenarios, it fails to make the correct association between detected and tracked objects. It is because of the noisy detection results and an inaccurate motion model in tracking. To improve data association, some data-driven deep learning methods have been proposed \cite{DBLP:journals/corr/abs-1905-02843, Milan:2017:AAAI_RNNTracking, learning_to_track, DBLP:journals/corr/abs-1806-11534, Giancola_2019_CVPR}, Baser et al. \cite{DBLP:journals/corr/abs-1905-02843} designed similarity and data association networks to solve the problem of data association. The similarity network, based on a Siamese neural network, processed the appearance and the 3D geometrical size and location to produce a similarity map. Then they performed the CNN-based data association on the map to identify associated pairs. Xiang et al. \cite{learning_to_track} formulated the MOT as decision-making in Markov Decision Processes (MDP), and they transformed the learning of similarity function for data association into policy learning for MDP in a reinforcement learning fashion. However, tracking methods based on deep learning require large tracking datasets with detailed annotations, and robots would also be equipped with a powerful computation unit, such as GPU, to perform properly in real-time.

We aimed to investigate how a tracker can improve the detector performance, rather than inventing a new method of tracking. Therefore, we chose the classical and efficient EKF to do the state estimation and the Hungarian algorithm to do the data association. The expected efficiency can allow our MODT framework to be generalized for small robot platforms.

\section{Detection and Tracking Approach}\label{detector_trackor}

\subsection{Detection Method}

\begin{figure}
    \centering
    \includegraphics[width=.45\textwidth]{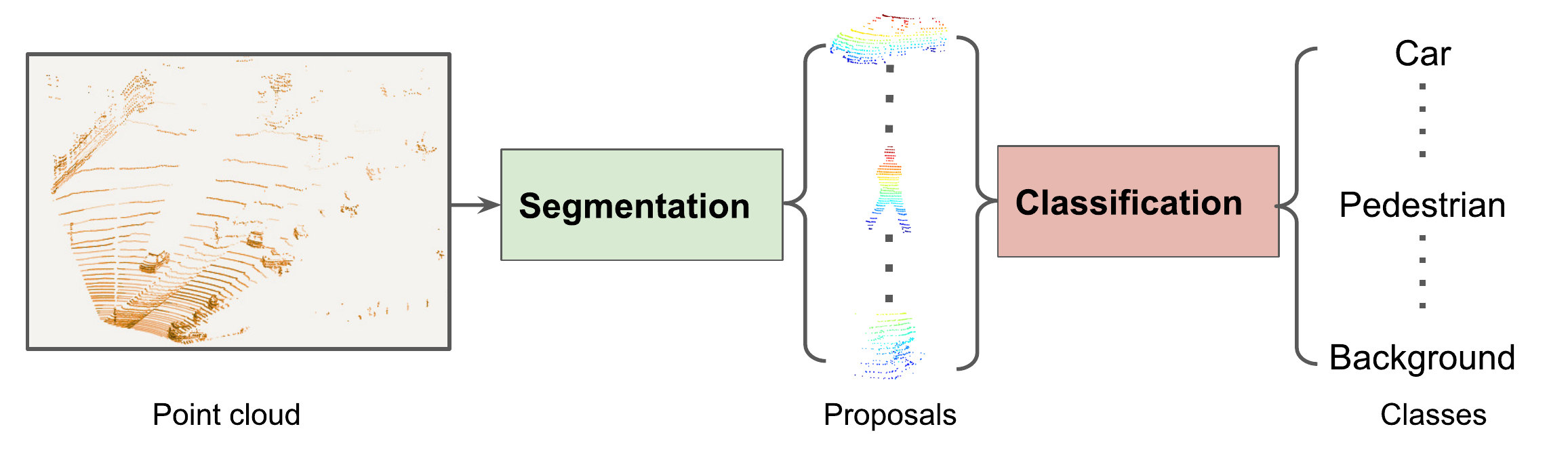}
    \caption{The pipeline of detection method.}
    \label{fig:det_method}
\end{figure}

An efficient two-stage detection method \cite{my_efficient_paper} is chosen for detecting 3D objects in our framework. It has two main components: point cloud segmentation and proposal classification, as shown in Fig. \ref{fig:det_method}. On the left, the class-agnostic segmentation module generates 3D proposals. The first step in segmentation is ground point removal, in which piece-wise multi constant planes are used to model the ground. Points that are below or close to the ground plane are discarded, and the remaining points are clustering into different 3D segments according to the Euclidean distance between each other. Each segment is considered a potential proposal, except for some improper ones, which are too large or too small. More technical details of point cloud segmentation can be found in \cite{my_efficient_paper}. It takes just 31 ms for this two-stage detection method to process one point cloud for generating 3D proposals.

For each 3D proposal, the size and location are already known, but its class is not. The aim of proposal classification is to identify the class of proposal. Every proposal includes about $400$ points. PointNet \cite{DBLP:journals/corr/QiSMG16}, which comprises a $1\times1\times1$ convolutional kernel in 3D, is used to perform the classification task for each proposal. Even though PointNet uses only a small kernel, its structure of multiple layers still requires around 260 ms to complete the classification task for all the proposals in one point frame on a standard CPU platform. It is clear that the proposed classification stage needs much more computation cost than point cloud segmentation. This makes it a potential bottleneck for real-time detection by robots with limited processing resources. This computation cost can be significantly reduced, however, if the classification model is launched at a low frequency and for just a few unknown proposals.

\subsection{Tracking Approach}
\subsubsection{Tracking Filter}
An EKF \cite{kalman_filter} is used to estimate the state for each object. The state is modelled as $\textbf{X}^{T}_{i|i} = [x_{i},y_{i},\theta_{i}, v_{i}]$ including the 2D location and velocity vector of the object. The height $z$ and the velocity in a vertical direction is not contained in $\textbf{X}^{T}_{i|i}$, because they are almost constant and a 2D plane is enough to describe the trajectory for tracked objects. The nominal motion model $F(\textbf{X}^{T}_{i|i})$ of inter-frame displacement for the tracked object is approximated by assuming a constant velocity, as in Equ. \ref{equ:motion_model}.
\begin{equation}
\begin{aligned}
\textbf{X}^{T}_{i+1|i} = F(\textbf{X}^{T}_{i|i}) = \begin{bmatrix}
           x_{i} + \delta t\times cos(\theta_{i})\times v_{i}\\
           y_{i} + \delta t\times sin(\theta_{i})\times v_{i}\\
           \theta_{i} \\
           v_{i}
         \end{bmatrix}
\label{equ:motion_model}
\end{aligned}
\end{equation}

$\textbf{X}^{T}_{i+1|i}$ is the predicted state after propagation to the next frame $i+1$, and $\delta t$ is the time interval between two contiguous 3D frames. The observation generated by the detection is $\textbf{O}_{i+1}=[\hat{x}_{i+1}, \hat{y}_{i+1}]$. The size and direction of the detected objects are generated by fitting a minimum bounding box to all points in one proposal instead of using regression network. Since the bounding boxes can be inaccurate, we do not put them in the observation vector for updating the state.

\subsubsection{Data Association}
To assign detected objects to existing objects, we define the assignment cost $C_{ij}$ of a matched pair (i,j) as the sum of three terms, i.e $C_{ij} = \alpha\times (1-I_{ij}) - \beta\times N_{ij} - \gamma\times S_{ij}$. $I_{ij}$, $N_{ij}$, and $S_{ij}$ represent the intersection-over-union, difference of point number, and difference of size between detection $i$ and existing object $j$, respectively, and $\alpha, \beta$, and $\gamma$ are their weighting factors. Many tracking algorithms \cite{baseline_tracking, SORT} calculate only intersection-over-union as the assignment cost. However, our multi-term cost can easily resolve ambiguity between traffic participants and background objects, especially when they are close to each other, such as a pedestrian walking past a pole. The Hungarian algorithm \cite{Kuhn55thehungarian} is used to calculate the optimal assignment based on the assignment cost matrix. A maximum assignment cost is imposed to reject associations whose cost is above the defined threshold of $T_{DA}$.


\section{Efficient Detection with Tracking}\label{efficient_detection}

\begin{figure}[htbp]
    \subfigure[Common pipeline of object detection and tracking]{
    \centering
    \includegraphics[width=.4\textwidth]{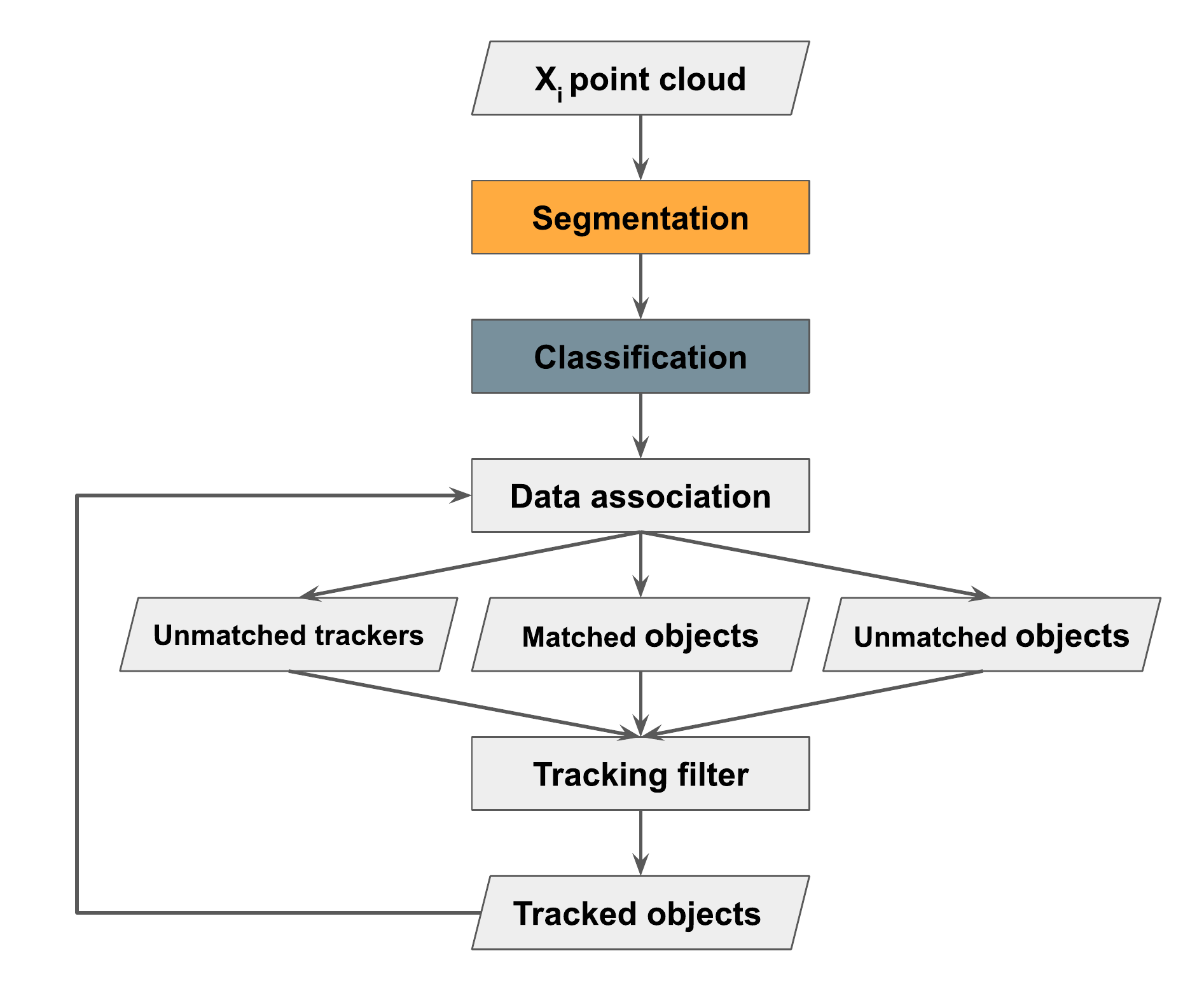}
    }
    
    \subfigure[Proposed pipeline of object detection and tracking]{
     \centering
     \includegraphics[width=.4\textwidth]{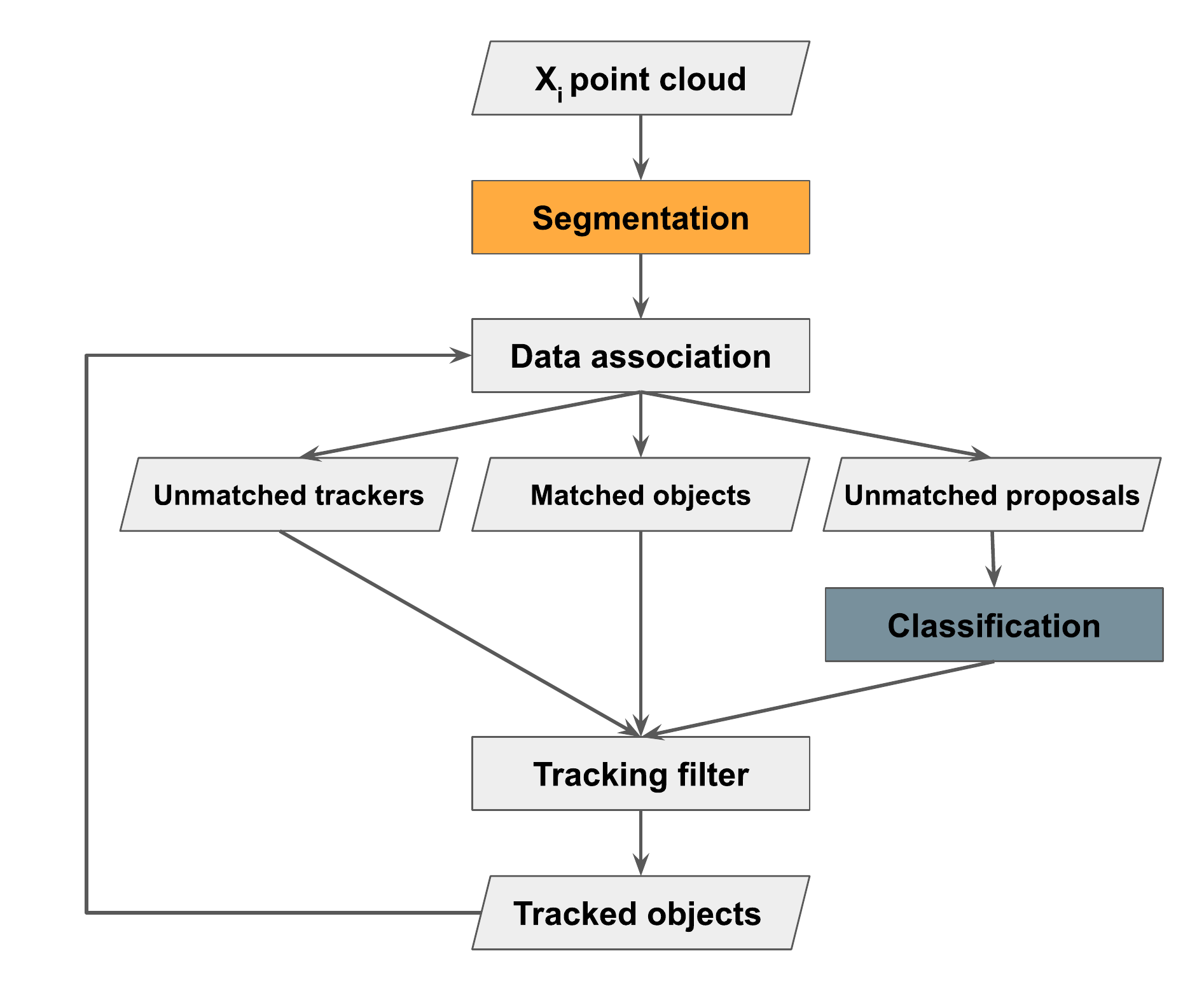}
    }
    \caption{Comparison of conventional and proposed MODT framework}
    \label{fig:framework}
\end{figure}
 
 The tracker can track objects moving at different velocities and various locations, but the class labels usually stay the same. These consistent class labels can be propagated to the proposals of the next point cloud frame using a prediction step and data association. The detection module does not need to perform classification again on these proposals with already-known class labels. This is the main idea for designing efficient detection with tracking. In this approach, the tracker propagates class information from previous detection results to the current frame and guides the detection module to selectively perform classification tasks.

\subsection{Methodology}

\begin{figure}
    \centering
    \includegraphics[width=.4\textwidth]{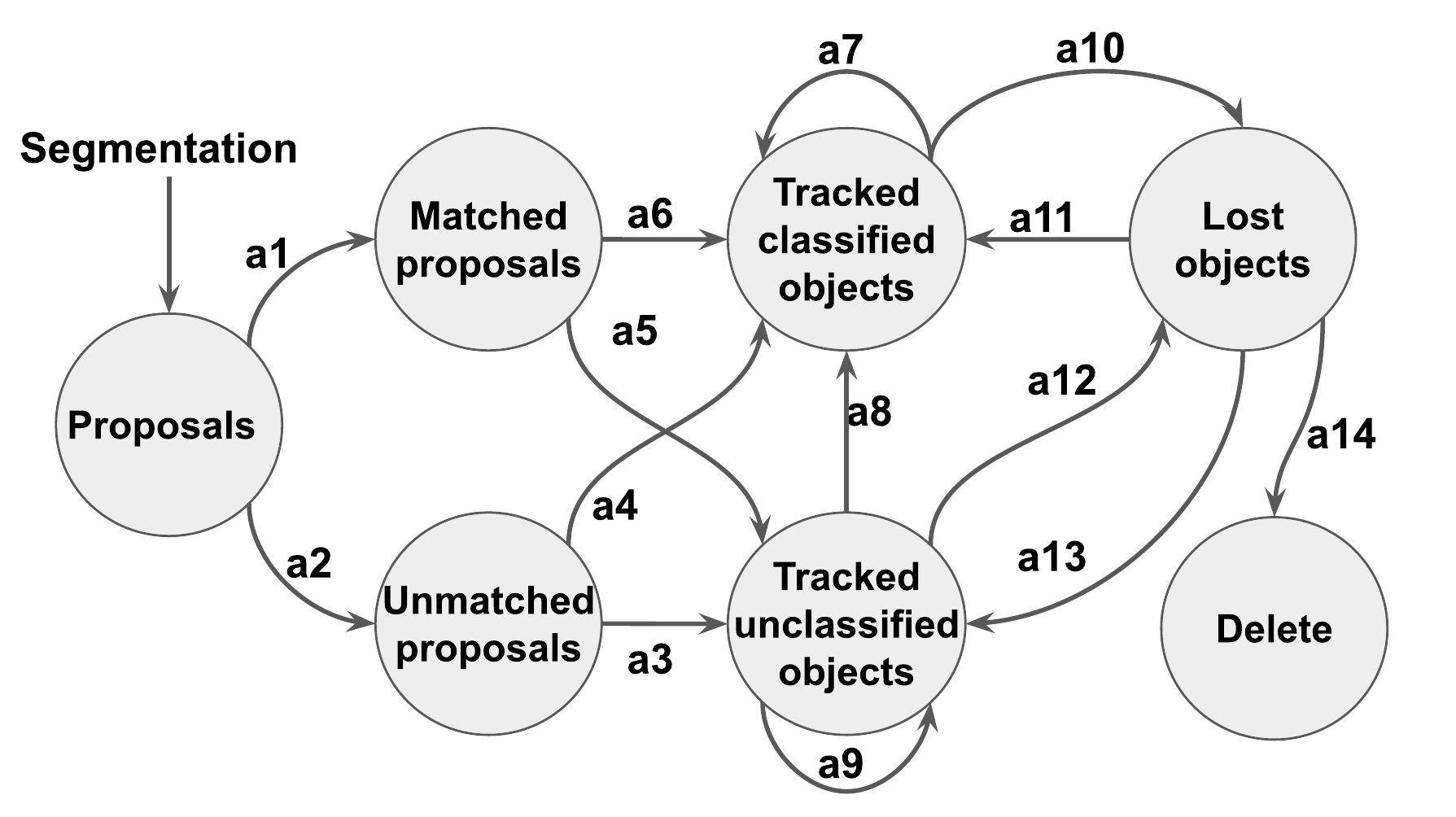}
    \caption{The finite state machine diagram for efficient detection.}
    \label{fig:fsm}
\end{figure}

The flow chart of our efficient detection method is depicted in Fig. \ref{fig:framework}. The point cloud $X_{i}$ is first passed into the segmentation algorithm, and $N$ proposals, $P=\{ p_{i}=[x_{i},y_{i},z_{i},l_{i},w_{i},h_{i},\theta_{i}]| 0\leq i \leq N\}$, are obtained. The conventional detection and tracking method, as shown in Fig. \ref{fig:framework}(a), performs classification for all proposals $P$ to generate objects with the class, $O=\{ o_{i}=[x_{i},y_{i},z_{i},l_{i},w_{i},h_{i},\theta_{i}, c_{i}]|0\leq i \leq N\}$, and class labels include background, car, pedestrian, and cyclist. Then, these objects $O$ go through data association to connect them with some trackers, $T=\{ t_{i}=[x_{i},y_{i},z_{i},l_{i},w_{i},h_{i},\theta_{i}, c_{i},v_{i}]|0\leq i \leq M\}$. We find that tracker $t_{i}$ and its associated object $o_{i}$ contain redundant information, class label $c_{i}$, while the computation of running the classification model for $c_{i}$ is quite expensive. Therefore, we proposed an efficient detection and tracking framework, as shown in Fig. \ref{fig:framework}(b), to reduce the computation of redundant information.

Data association is executed on all proposals before classification, and $P=\{p^{match}_{i},p^{unmatch}_{j}| 0\leq i\leq N_{1}, 0\leq j\leq N_{2}\}$. The proposals $P^{match} = \{p^{match}_{i}| 0\leq i\leq N_{1} \}$ associated with known trackers are assigned identical class labels. The expensive classification model is applied only to the unmatched proposals $P^{unmatch} = \{p^{unmatch}_{j}| 0\leq j\leq N_{2} \}$. Since the numbers of proposals and objects are the same, the data association computation of our method is the same as the conventional one. The new object gradually enters the field of view and the unmatched proposals $N_{2}$ is usually much fewer than matched objects $N_{1}$. Therefore, our framework can significantly reduce the running time of the classification model. 

\begin{table}[]
\caption{Detailed description of input for transition.}
\label{tab:input}
\begin{tabular}{ m{0.8cm}| m{6cm} }
 \hline
 \textbf{Input} & \textbf{Description} \\
 \hline
 a1 & Proposals are associated with existing tracked objects \\
 \hline
 a2 & Proposals are not associated with any tracked objects \\
\hline
 a3 & The classification method is executed to generate class labels for these unmatched proposals \\
\hline
 a4 & Associated proposals are used to update the location of tracked objects \\
\hline
 a5 & Matched proposals are assigned the same corresponding class label \\
 \hline
 a6 & The tracked objects are not associated with any proposals \\
\hline
 a7  & The lost objects are re-associated with proposals \\
\hline
 a8  & The lost objects are not associated with any proposals \\
\hline
 a9  & The lost objects are not associated with any proposals for $N$ consecutive times  \\
\hline
\end{tabular}
\end{table}

\subsection{Finite-state Machine for Every Proposal}

All possible states for every proposal entering the region of interest are depicted in Fig. \ref{fig:fsm}. As indicated earlier, proposals are generated by the segmentation process and they are switched to the state of being matched or unmatched through data association. Regardless of its matching state, every object is tracked by the tracker. The tracked objects are switched to lost objects if they fail to be associated with any current proposals, otherwise, they continue to be the tracked objects. Lost objects that have not been associated with any proposals for a certain number of consecutive frames are deleted from the tracking process. They tend to be objects outside the region of interest rather than temporally occluded objects. If they are successfully associated with proposals, they are switched back into the tracked objects. All the input to the finite-state machine can be found in Table \ref{tab:input}.

\begin{figure}[b]
\begin{adjustwidth}{0 cm}{}
    \hspace{0cm}
    \includegraphics[width=.4\textwidth]{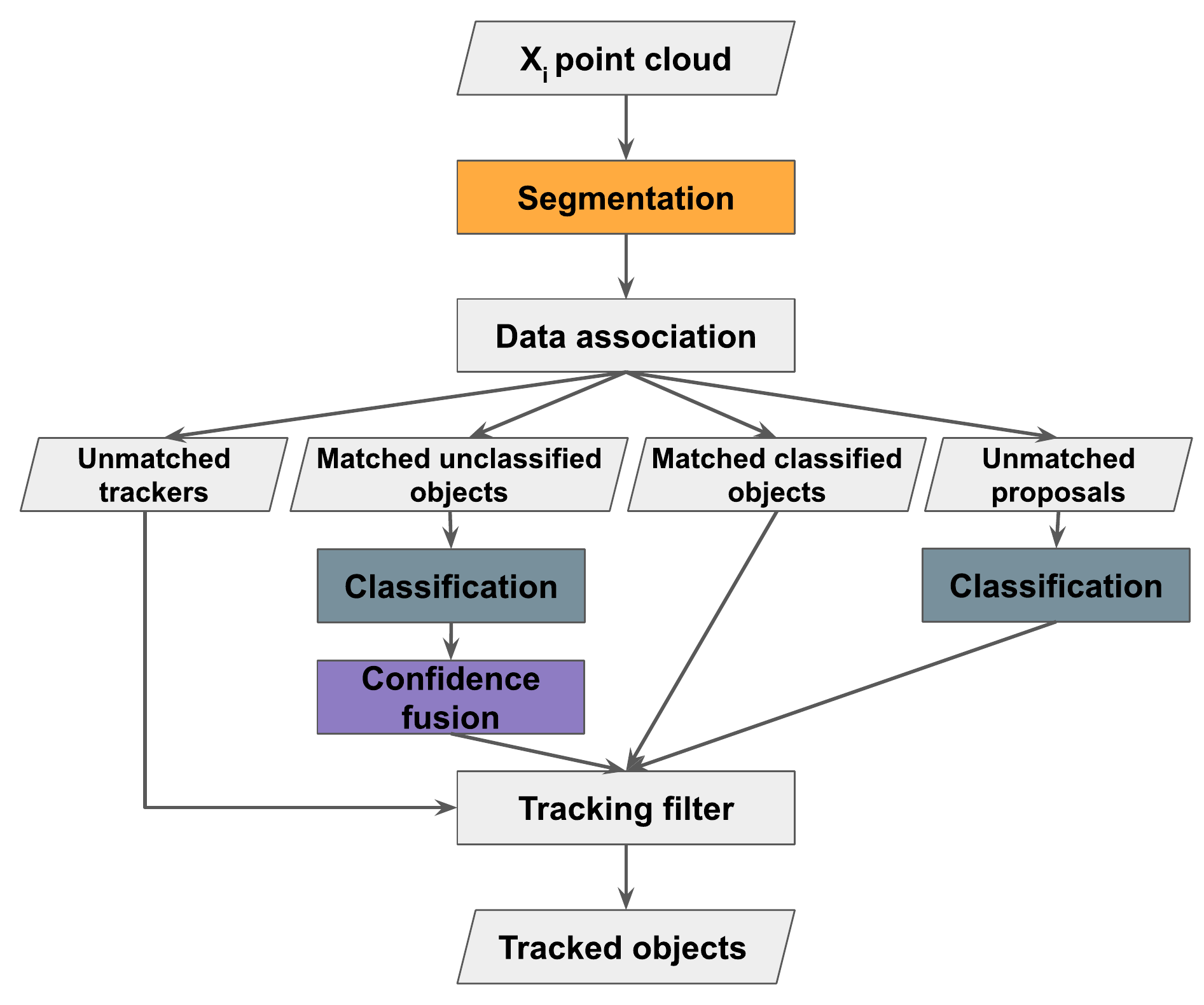}
    \caption{Framework for accuracy detection with tracking}
    \label{fig:acc_track}
\end{adjustwidth}
\end{figure}

\begin{figure}
    \centering
    \includegraphics[width=.4\textwidth]{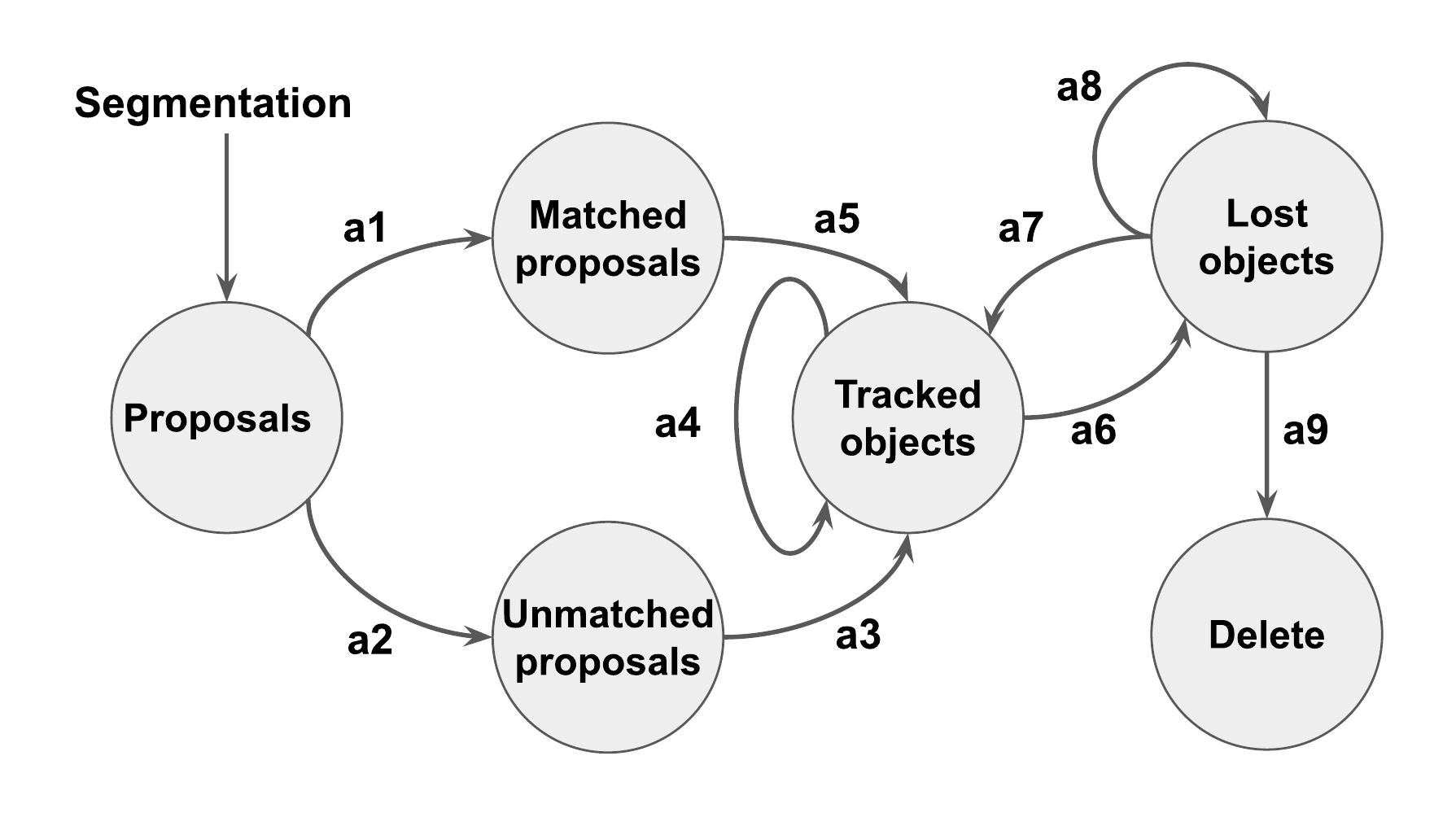}
    \caption{The finite-support machine.}
    \label{fig:fsm_2}
\end{figure}

\subsection{Theoretical Analysis of Energy Saving}
It is assumed that there are on average $N_{obj}$ foreground objects and $N_{bg}$ background objects in one frame of a point cloud. The average number of temporal frames that foreground and background objects will travel through is $N_{go}$. Instead of using an algorithm for time complexity, we adopt energy complexity \cite{energy_complexity_model} to compare the energy consumption of different algorithms. The energy complexity of segmentation for one point cloud is $E(seg)$, and the classification of one object is $E(class)$. For a tracking task with $M$ frames of the point cloud, the energy consumption of standard tracking-by-detection is expressed in Equ. \ref{equ:standard_tracker} $E_{1}(M)$. The energy consumption for the hybrid efficient tracking is $E_{2}(M)$ in Equ. \ref{equ:standard_tracker}. The ratio is approximately $N_{go}$ (see Equ. \ref{equ:ratio}). As more objects traverse the point cloud, our method saves more energy. Object detection is one of the most important tasks in the perception system, and it is launched when the machine is turned on, so $N_{go}$ is usually large. Hence, efficient tracking and detection can greatly decrease energy consumption.

\begin{equation}
\begin{aligned}
&E_{1}(M) = M \times E(seg) + M \times (N_{obj}+N_{bg}) \times E(class) \\
&E_{2}(M) = M \times E(seg) + \frac{M}{N_{go}} \times (N_{obj}+N_{bg}) \times E(class)
\label{equ:standard_tracker}
\end{aligned}
\end{equation}

\begin{equation}
\begin{aligned}
\gamma & = \frac{E_{1}(M)}{E_{2}(M)} = \frac{E(seg) + (N_{obj}+N_{bg}) \times E_(class)}{E(seg) + \frac{1}{N_{go}} \times (N_{obj}+N_{bg}) \times E(class)} \\ 
&=\frac{1 + (N_{obj}+N_{bg}) \times a}{1 + \frac{1}{N_{go}} \times (N_{obj}+N_{bg}) \times a} \\ 
&= \frac{1 +  \frac{1}{(N_{obj}+N_{bg}) \times a} }{ \frac{1}{(N_{obj}+N_{bg}) \times a} + \frac{1}{N_{go}}} \approx N_{go}
\label{equ:ratio}
\end{aligned}
\end{equation}

where, $a=\frac{E(class)}{ E(seg)}$ and $(N_{obj}+N_{bg})\times a \gg 1$, so  $\frac{1}{(N_{obj}+N_{bg}) \times a} \approx 0$

\section{Accurate Detection with Tracking}\label{accurate_detection}

The tracker can keep following the objects and provide a sequence of their different views. These views enable higher classification confidence for the objects. For example, as human beings, we keep observing ambiguous objects and actively or passively collect different views and information about them until we recognize them with certainty. To design a mechanism to increase detection confidence, we must define the independence about different views of the same object. Therefore, the classification model must classify the object with highly different views, and the classification results are fused by a probabilistic model.

\begin{table}
\caption{Detailed description of input for transition.}
\label{tab:acc_det}
\begin{tabular}{ m{0.8cm}| m{7cm} }
 \hline
 \textbf{Input} & \textbf{Description} \\
 \hline
 a1 & Proposals are associated with existing tracked objects \\
 \hline
 a2 & Proposals are not associated with any tracked objects \\
\hline
 a3 & The classification method is executed and return a uncertain classification confidence \\
\hline
 a4 & The classification method is executed and return a distinct classification confidence \\
\hline
 a5 & Proposals are associated with unclassified objects \\
 \hline
 a6 & Proposals are associated with classified objects \\
\hline
 a7  & Associated proposals are used to update the location of tracked certain objects \\
\hline
 a8  & Confidence fusion model returns a high classification confidence \\
\hline
 a9  & Associated proposals are used to update the location of tracked unclassified objects \\
\hline
 a10 & Tracked certain objects are not associated with any proposals \\
\hline
 a11  & Lost classified objects are re-associated with proposals \\
\hline
 a12  & Tracked classified objects are not associated with any proposals \\
\hline
 a13  & Lost unclassified objects are re-associated with proposals \\
\hline
 a14  & Lost objects are not associated with any proposals for $N$ consecutive times  \\
\hline
\end{tabular}
\end{table}

\subsection{Methodology}
The framework of detection with tracking is shown in Fig. \ref{fig:acc_track}. According to the classification confidence, the unmatched proposals are divided into two groups: classified objects and unclassified objects. The classified objects usually present an explicit view and feature to the classification model, and they receive very high classification confidence. However, unclassified objects could be far away or they are occluded by other objects, hence, only a portion of them is captured by the sensor. In this case, the classification model would fail to return a distinguishable classification with confidence. For tracked and unclassified objects, the classification model is applied to the corresponding proposals for further classification. If the fusion confidence is higher than a user-defined threshold, the tracked object is switched into a classified object.

\subsection{Finite-state Machine for Every Proposal}

The FSM for every proposal in accurate detection with tracking is depicted in Fig. \ref{fig:fsm_2}, and the corresponding inputs for transition are shown in Table \ref{tab:acc_det}. Compared with the FSM for efficient detection, the FSM for accurate detection divides the state of tracked objects into two sub-categories: classified objects and unclassified objects. When the object is in the state of tracked unclassified objects, the classification model and confidence fusion model are launched to re-adjust the detection confidence until the confidence about a class is high enough.

\subsection{Confidence Fusion}

\begin{figure}
    \centering
    \includegraphics[width=.45\textwidth]{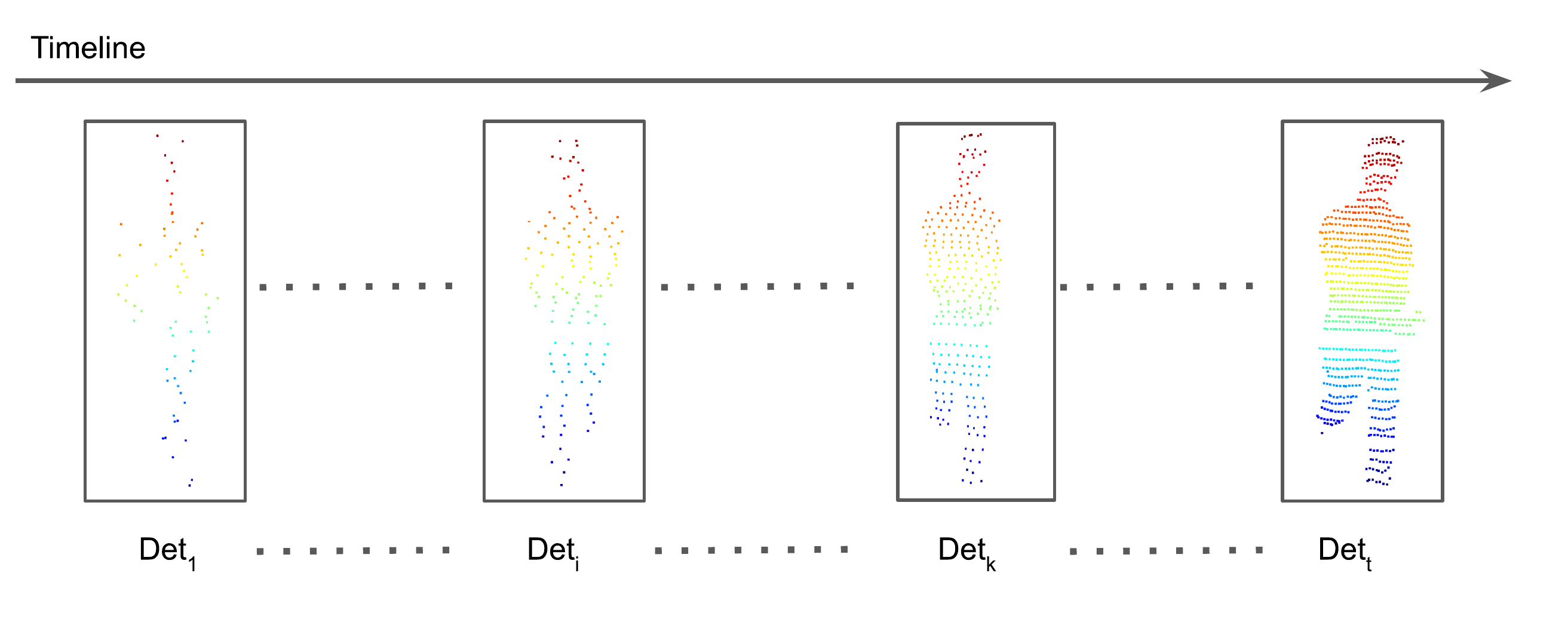}
    \caption{Point cloud stream. The point cloud shows views about the same object in different time indexes.}
    \label{fig:pc_stream}
\end{figure}

The tracker can provide a stream of 3D segments about one tracked object, which could be the background, a car, a pedestrian, or a cyclist, see Fig. \ref{fig:pc_stream}. The stream is denoted as $Det_{1:t}$, in which $Det_{i}$ is the point cloud $M\times 3$ for one tracked object. Its class is represented with a discrete random variable $X$ with a $4\times 1$ probabilistic mask. After it processed the point cloud $Det_{i}$, the classification model $G$ returns the observation of $X$, which is also denoted as a discrete random variable $Y_{i}$. The $Y_{i}$ is implemented by a $4\times 1$ probabilistic mask. Therefore, we have $P(Y_{i}|X)= G(Det_{i})$, which can be treated as the likelihood of observation $Y_{i}$.

\begin{figure}
    \centering
    \includegraphics[width=.45\textwidth]{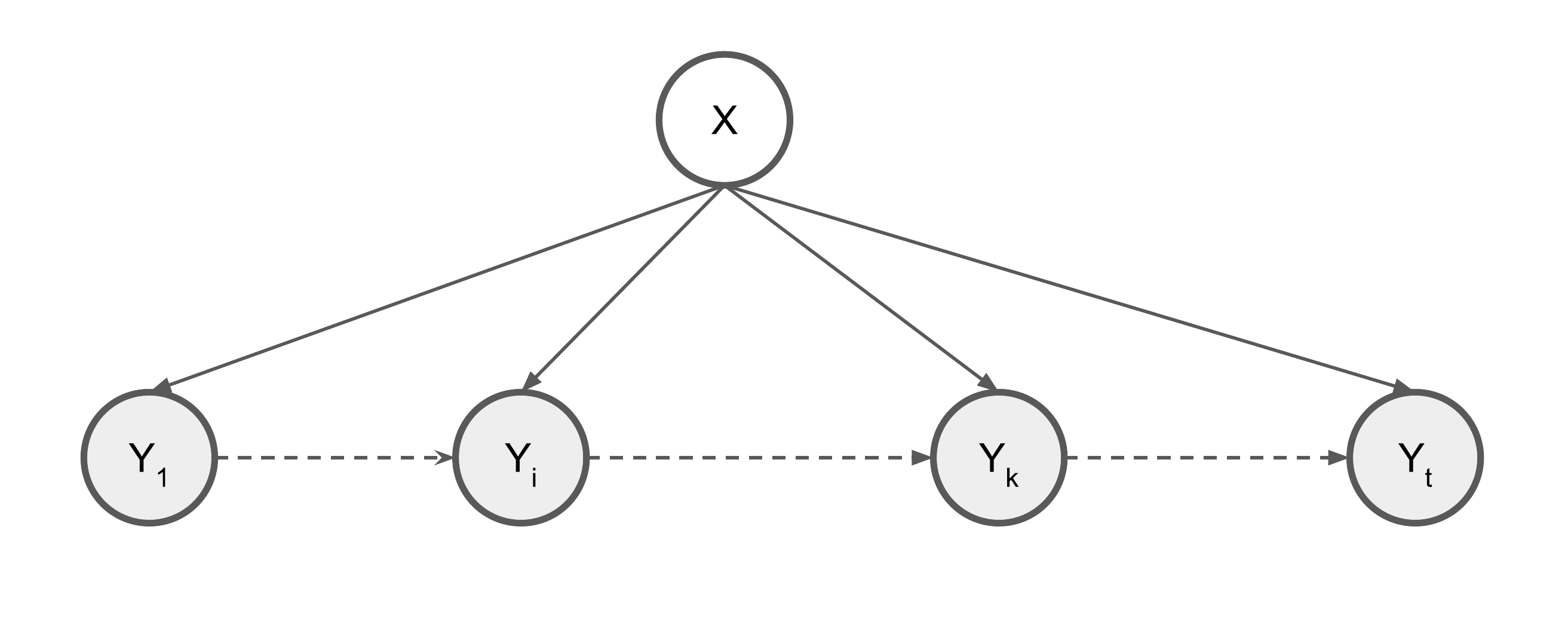}
    \caption{Probabilistic graph model.}
    \label{fig:pgm}
\end{figure}

The objective is to fuse all $Y_{1:t}$ together to calculate the confidence about $X$, i.e. the estimation of $P(X|Y_{1:t})$. The probabilistic graphical model of our fusion model is shown in Fig. \ref{fig:pgm}. According to the Bayes theorem \cite{Bayes_63} and Markov property, $P(X|Y_{1:t})$ can be expanded as Equ. \ref{equ:full_ex}.

\begin{equation}
\begin{aligned}
& P(X|Y_{1:t}) = \frac{P(Y_{1:t}|X)P(X)}{P(Y_{1:t})} \propto P(Y_{1:t}|X)P(X) \\ & = P(Y_{t}|Y_{1:t-1},X)P(Y_{t-1}|Y_{1:t-2},X)......P(Y_{1}|X)P(X)
\end{aligned}
\label{equ:full_ex}
\end{equation}

One proposal normally has similar views, shapes, sizes, and other properties in two adjacent frames of the point cloud. Therefore, the classification model is very likely to return similar results for these two adjacent measurements. If both of them are used to update the classification confidence, this would make the classification overconfident. To make an inference on one proposal from different perspectives, it is assumed that $Y_{i}$ and $Y_{j}$ are independent observations if the corresponding point cloud $Det_{i}$ and $Det_{j}$ are sufficiently different from each other. Otherwise, $Y_{i}$ and $Y_{j}$ are treated as the same observation. The difference between $Det_{i}$ and $Det_{j}$ relies on their geometrical shape and the number of points. A user-defined threshold is used to determine whether they are independent observations or not. Based on these assumptions, we get Equ. \ref{equ:indedpend}, and Equ. \ref{equ:full_ex} can be further simplified into Equ. \ref{equ:full_inden_dep}.

\begin{equation}
\hspace{-0.0cm}
P(Y_{j}|Y_{i},X)=\begin{cases}
    P(Y_{j}|X), & \text{if $Y_{j}$ is independent of $Y_{i}$}.\\
    1, & \text{Otherwise}.
    \label{equ:indedpend}
  \end{cases}
\end{equation}

\begin{equation}
\begin{aligned}
&  P(X|Y_{1:t}) \propto P(Y_{1:t}|X)P(X) \\ & = P(Y_{n}|X)....P(Y_{k}|X)...P(Y_{1}|X)P(X)
\\ & = \prod_{j=1}^{m}P(Y_{I(j)}|X)P(X) = P(Y_{I(m)}|X)\prod_{j=1}^{m-1}P(Y_{I(j)}|X)P(X)
\\ & = P(Y_{n}|X)\prod_{j=1}^{m-1}P(Y_{I(j)}|X)P(X)
 \propto P(Y_{n}|X)P(X|Y_{1:I(m-1)})
\end{aligned}
\label{equ:full_inden_dep}
\end{equation}

where independent frames $I = \{1,...,k,....,n\}$ are selected from all $t$ frames, and the number of selected frames is $m$, ($n \leq t$, $m \leq t$). $P(X|Y_{1:I(m-1)})$ carries all the past classification information and describes the fusion result of a tracker at frame $I(m-1)$.

The classification model is launched only when a new independent observation $Y_{j}$ is received. Once the tracker is confident enough about one class, the classification confidence is fixed, and the proposal is tracked without performing additional classification.

\section{Experimental Results}\label{experiments_part}

\subsection{Implementation Details}\label{Implementation} 
\subsubsection{Dataset}
The KITTI benchmark dataset \cite{geiger2012we} is adopted to validate our proposed method. It includes several episodes of videos with the corresponding 3D LIDAR point cloud, RGB image, and calibration parameters. Some episodes also include annotations of detection and tracking tasks. Besides, different scenarios such as city, campus, and residence, also are covered in the dataset. The episodes of videos 0001, 0005, 0014, 0019, 0020, 0035, 0059, and 0084, are selected from the raw dataset in KITTI \cite{geiger2012we} for these experiments, as all of them include annotation data and multiple scenarios.

\subsubsection{Detection and Tracking Method Details}
For point cloud segmentation, we chose the same set of parameters for ground removal and segmentation as described in \cite{my_efficient_paper}. The structure of the classification model also is similar, but the size of the feature channel is decreased by half to simulate scenarios where the real detector cannot satisfactorily perform the detection. The input for the classification model is $400 \times 3$, where 400 points are input and there are 3 features for each point. The sizes of the corresponding feature maps in the classification model are $400 \times 32$, $400 \times 64$, and $400 \times 512$. After the max-pooling operation, $3$ layers of fully connected network compress the feature ($1\times 512$ to $1\times 4$) through two middle features ($1 \times 256$ and $1 \times 128$).

The processors in our hardware platform are eight Intel(R) Xeon(R) CPU E5-1620 v3 @ 3.50 GHz, and every core has two threads. The memory capacity is 16 Gigabytes. To make our method applicable to robots with limited processing capabilities, we set an affinity mask to restrict our code to one core when generating the experimental data. This helps to ensure that our method can run smoothly on small or medium robot platforms. Based on this software and hardware configuration, it takes 260 ms to perform the classification task for one frame of the point cloud.

The training procedures usually require 100 epochs to reach an adequate loss value \cite{my_efficient_paper}, which results in very high accuracy and close to that of an ideal detector. Since the classification performance is already good enough, the improvement in accuracy made by the confusion model in the tracker is marginal. Hence, it is difficult to assess the performance of the fusion model. Therefore, we stop the training model around 40 epochs to simulate the case in which detection is not accurate because of the limited number of training datasets or complicated scenarios. Apart from that, we set up the experiment with an ideal detector and tracker to validate the case in which detection is performed with a credible classification model.

As for the parameters in data association, $\alpha$, $\beta$, and $\gamma$ are set to 2, 0.01, and 0.1, and the rejection threshold $T_{DA}$ is 0.95. For the tracker, the states of velocity and direction are initialized to zero.

\begin{figure}
    \centering
    \includegraphics[width=0.4\textwidth]{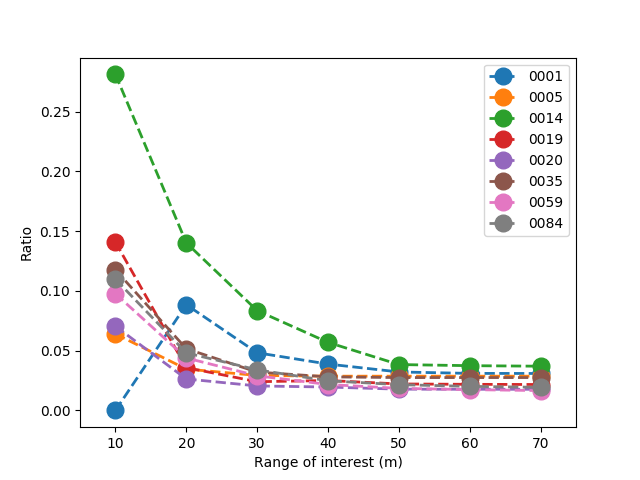}
    \caption{Efficiency against the distance of interest with the ideal detector and tracker.}
    \label{fig:efficiency}
\end{figure}

\subsubsection{Evaluation metrics}
The efficiency and accuracy of the detection method with and without the tracking function were compared. The improvement in efficiency is the ratio $\beta$ of the number of running classification models in the proposed detection framework $N_{p}$ to those in the conventional detection framework $N_{c}$, hence, $\beta ={N_{p}}/{N_{c}}$.

The detection performance is measured with precision-recall curves (PRC), which are the main metric for various detection benchmarks. A PRC can illustrate the precision $\theta$ concering different recall values $\gamma$. $\theta = TP/(TP + FP)$ and $\gamma = TP/(TP + FN)$, where $TP$ is the number of true positives (correctly segmented objects), $FN$ is the number of false negatives (missed objects), and $FP$ is the number of false positives, i.e. the incorrectly detected objects.

\begin{figure*}[ht]
\centering
\begin{adjustwidth}{-1.0 cm}{}
    \hspace{0pt}%
    \subfigure[Cars.]{
     \includegraphics[width=.35\textwidth]{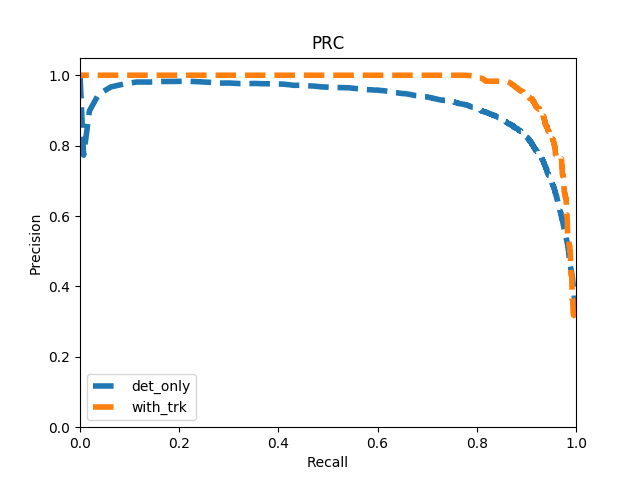}}
    \hspace{0pt}%
    \subfigure[Cyclists.]{
    \includegraphics[width=.35\textwidth]{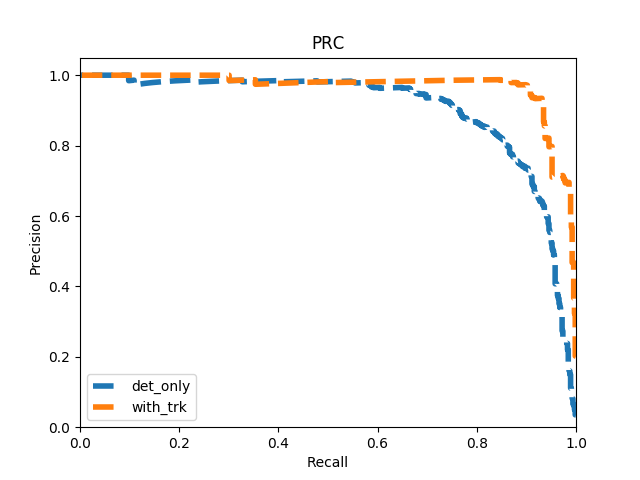}}
    \hspace{0pt}%
    \subfigure[Pedestrians.]{
    \includegraphics[width=.35\textwidth]{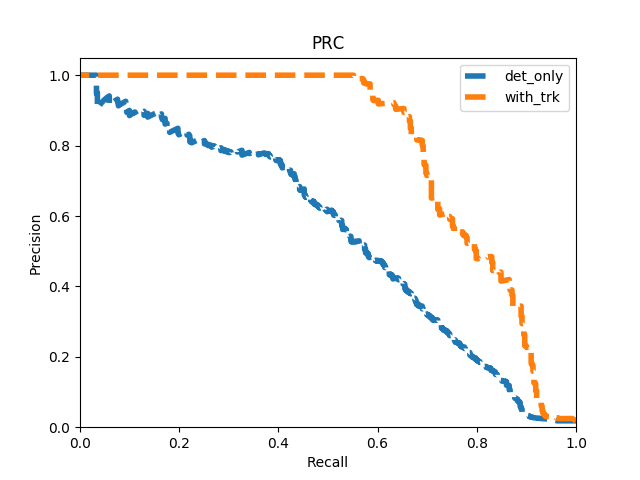}}
\end{adjustwidth}
\caption{PRCs for experiments with the real detector and an ideal tracker. The curve of $det\_only$ represents the detection results without running the tracking function, while $with\_trk$ includes the tracking function.}
\label{fig:good_model}
\end{figure*}

\subsection{Experiments with Ideal Detector and Tracker}

\begin{table}[h!]
\caption{The average number $N_{go}$ in different datasets.}
\label{tab:streak}
\centering
    \begin{tabular}{c|c|c|c|c|c|c|c|c}
    \hline
    Dataset&0001&0005&0014&0019&0020&0035&0059&0084\\
    \hline
    $N_{go}$& 32.4 & 34.9 & 27.1 & 46.3 & 56.6 & 36.4 & 61.9 & 51.7  \\
    \hline
    \end{tabular}
\end{table}

To validate the best performance that our proposed framework could achieve, we assumed that both the detector and the tracker used in the framework are ideal. That is, the detector can segment all proposals and recognize perfectly the class of objects in the interest range, and the tracker can complete the data association and state estimation without error. To simulate the ideal detector and tracker, we adopt directly the ground truth data of detection and tracking. Since the ideal detector produces 100\% correct detection results, detection with or without the tracking function will achieve the same performance. Therefore, we only investigate how efficiency is improved with our proposed framework. 

The whole point cloud is first processed by the segmentation module to extract the proposals, and the classification model is launched only for proposals within the range of interest. Once a proposal is placed by the classification model into a class, it is labelled as an object, and one tracker is assigned to keep tracking it without performing classification on it. The range of interest depends on the application domain. If robots move fast and predictions of future abnormal events are vital, such as with autonomous vehicles, the level of interest should be very large. If the robot is always at low speed and faraway objects are not important, the range of interest could be set narrow to save computation resources. The efficiency improvement is shown in Fig. \ref{fig:efficiency}. It shows that when the range of interest is set to cover the entire perception area (70 $m$), the proposed detection framework with a tracker can reduce the number of running classification models to only 2\% of the original full detection. Even with a short range of interest (10 $m$), the average ratio for efficiency improvement is still as low as 10\%. 

In Fig. \ref{fig:efficiency}, the ratio decreases with the range of interest, because the lifespan of a tracker increases as the range of interest becomes bigger. The longer the lifespan, the more efficiency is improved. The average lifespan of the tracker is shown in Table \ref{tab:streak}. It shows that the efficiency improvement $\beta$ is approximately equal $\frac{1}{N_{go}}$, as predicted in Equ. \ref{equ:ratio}. According to this experiment, it can also be concluded that our efficient detection framework with tracking could significantly reduce the computation effort.

\subsection{Experiments with the Real Detector and an Ideal Tracker}

\begin{table}[h]
\caption{Classification accuracy (mAP) for every object.}
\label{tab:good_model_map}
\centering
    \begin{tabular}{p{1.7cm}|p{1.7cm}|p{1.7cm}|p{1.7cm}}
    \hline
    Method  & Car(\%) & Cyclist(\%) & Pedestrian(\%) \\
    \hline
    det\_only  & 92.2 & 90.4 & 54.4 \\
    \hline
    with\_trk  & 97.4 & 96.8 & 79.3 \\
    \hline
    \multicolumn{4}{p{8.2cm}}{'det\_only' is detection without tracking; 'with\_trk' is one with tracking. }
    \end{tabular}
\end{table}

In this experiment, we ran the segmentation to extract all proposals and used the classification algorithm to predict the confidence level. Hence, the 3D bounding box with confidence for each object can be generated by the real detector. Each detected object is assigned a tracking ID according to the ground truth of the tracking dataset. The tracking ID guarantees a perfect data association, which can make state estimation much simpler and more accurate. Therefore, we call it the ideal tracker.


The point cloud segmentation and proposal classification models are run for every frame in the point stream and the detection results are stored. Meanwhile, detection with the tracking function and fusion model is implemented to process the same point stream, and the detection results updated by the trackers are also stored. The PRC and mean average precision (mAP) are calculated as indicators to compare the accuracy of two detection results. The efficiency of the real detection model is evaluated with $\beta$. Besides, we conducted a series of experiments to investigate and compare how parameters in a fusion model affect detection accuracy and efficiency.

\subsubsection{Detection Performance}
The mAP comparison of methods with and without tracking is shown in Table \ref{tab:good_model_map}. There is an obvious improvement in detection accuracy when tracking is used, and especially for detecting pedestrians. As for detecting cars, the detection model is already very accurate, $92.2\%$, so the tracker can yield very little improvement. However, when the training data are limited or the scenarios are difficult, the detection model does not reliably recognize a class of objects, such as pedestrians, the tracker shows an impressive improvement in detection accuracy, from $54.4\%$ to $79.3\%$. The PRC comparison is shown in Fig. \ref{fig:good_model}. It shows that, when the recall is small, detection by the two methods is almost equal and it shows little difference with and without the tracking function. As the recall increases and a large number of background objects are mis-recognized as the target foreground object, the method with a tracker shows greater precision than the one without a tracker. This means that the fusion model in the tracker can suppress the confidence in wrongly classified objects and boost the confidence in the target object.

\begin{figure}[h]
\centering
\begin{adjustwidth}{-0.0 cm}{}
     \includegraphics[width=.4\textwidth]{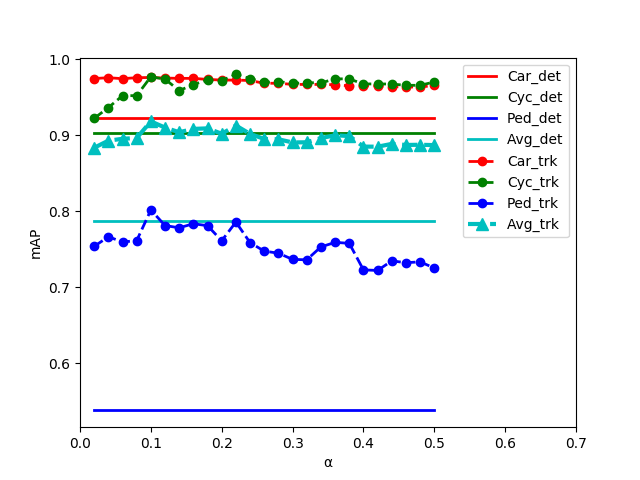}
\end{adjustwidth}
\caption{The mAP against different $\alpha$. 'Car\_det', 'Cyc\_det', 'Ped\_det', and 'Avg\_det' are the detection performance without tracking for car, cyclist, pedestrian, and average score. '*\_trk' denotes the detection performance with tracking.}
\label{fig:ratio_map}
\end{figure}

\subsubsection{Independence Assumption for Confidence Fusion}

\begin{figure}[h]
\centering
\begin{adjustwidth}{-0.0 cm}{}
     \includegraphics[width=.4\textwidth]{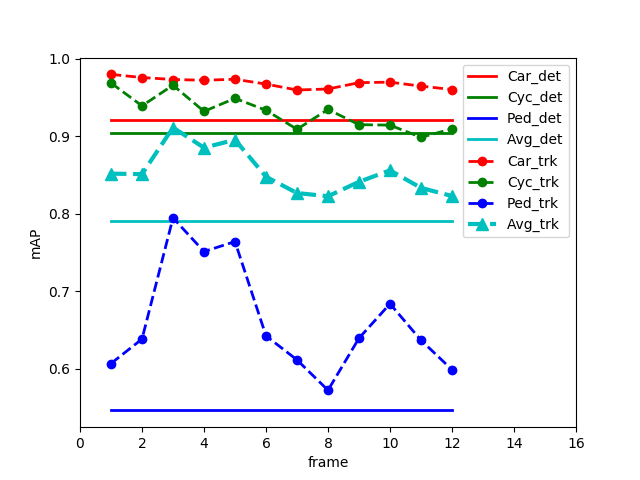}
\end{adjustwidth}
\caption{The mAP against different starting frames. The notation of lines is the same as in Fig. \ref{fig:ratio_map}.}
\label{fig:start_frame}
\end{figure}

The confidence fusion module is based on the assumption that the two different views are independent. To justify this assumption, we define a ratio of the number of points between two different views, i.e $\alpha = (N_{cur} - N_{pre})/N_{pre}$, where $N_{cur}$ is the number of the current point cloud, and $N_{pre}$ represents the number of a previously fused point cloud. By changing the $\alpha$, we can make our independence assumption strict or loose. A strict assumption would decrease the frequency of updating detection results with the fusion model. This could limit the noisy detection results in the early stage. By contrast, a loose assumption could increase the updating frequency and enable a fast convergence to the correct class when the detection result is not confident about one class. To investigate how the independence assumption, $\alpha$, affects the detection performance under tracking, we tested a set of $\alpha$ from 0 to 0.5 with an interval of 0.02. The results are displayed in Fig. \ref{fig:ratio_map}, which shows that the change of $mAP$ against $\alpha$ depends on the quality of the detector. For car detection, the detector can already produce high-quality detection results, and the independence assumption $\alpha$ does not have a noticeable impact on accuracy, and the accuracy against different $\alpha$ is almost constant. As for the cyclist and pedestrian, the change of accuracy for a pedestrian is much larger than that for a cyclist because of the quality of detection for these two classes. The optimal $\alpha$ also varies by object class, and the optimal $\alpha$ for a pedestrian ($0.16$) is smaller than that for a cyclist ($0.22$). One reason for this might be that the lower $\alpha$ had increased the updating frequency and increased the accuracy of the poor detector.

\begin{figure}
    \centering
\begin{adjustwidth}{-0.2 cm}{}
    \includegraphics[width=.4\textwidth]{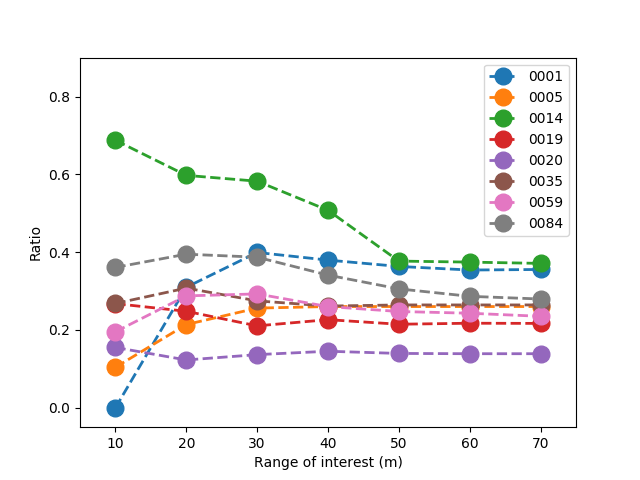}
    \caption{Efficiency against the distance of interest with the real detector and ideal tracker.}
    \label{fig:efficiency_fusion_ideal}
\end{adjustwidth}
\end{figure} 

\subsubsection{Starting Frame}
In the real experiment, we found that when tracking starts, the object is usually far away from the robot and the detection result is noisy and unreliable. To prevent a very noisy detection result from interfering the tracker, we launched the confusion model after tracking one object consecutively for a while. The performance of detection regarding different starting frame numbers can be found in Fig. \ref{fig:start_frame}. It shows that discarding the first three frames improves the detection results with tracking because the fusion model can combine more accurate detection results. However, as the number of frames increases, the improvement deteriorates. When the number of frames is close to the lifespan of a tracklet, detection with or without tracking it not important as chances of launching the confusion model decrease significantly.

\begin{figure}
    \centering
    \includegraphics[width=.4\textwidth]{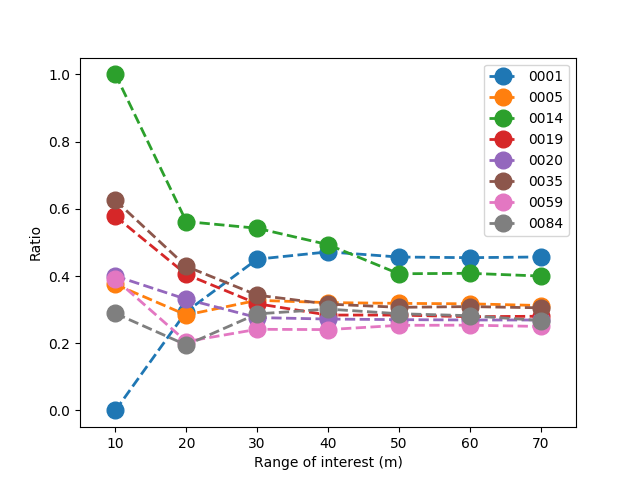}
    \caption{Efficiency against the distance with the real detector and tracker.}
    \label{fig:efficiency_real_tracker}
\end{figure} 

\begin{table}
\caption{Accuracy of classification (mAP) for every object.}
\label{tab:real_detector_tracker}
\centering
    \begin{tabular}{p{1.7cm}|p{1.7cm}|p{1.7cm}|p{1.7cm}}
    \hline
    Method  & Car(\%) & Cyclist(\%) & Pedestrian(\%) \\
    \hline
    det\_only  & 92.0 & 89.6 & 57.5 \\
    \hline
    with\_trk  & 93.8 & 91.2 & 73.0 \\
    \hline
    \end{tabular}
\end{table}

\begin{figure*}[t]
\centering
\begin{adjustwidth}{-1.0 cm}{}
    \hspace{0pt}%
    \subfigure[Cars.]{
     \includegraphics[width=.35\textwidth]{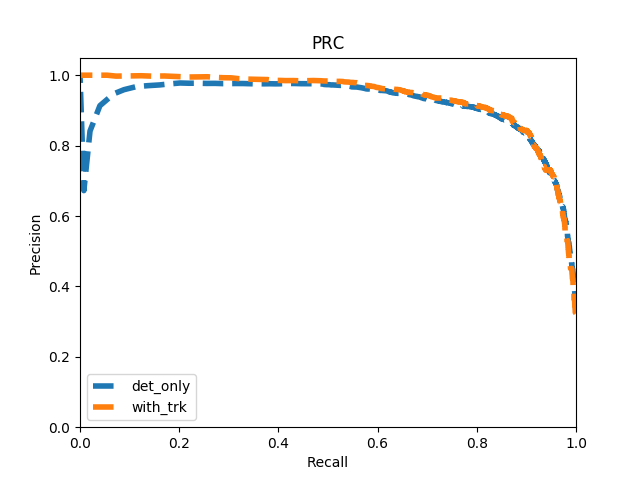}}
    \hspace{0pt}%
    \subfigure[Cyclists.]{
    \includegraphics[width=.35\textwidth]{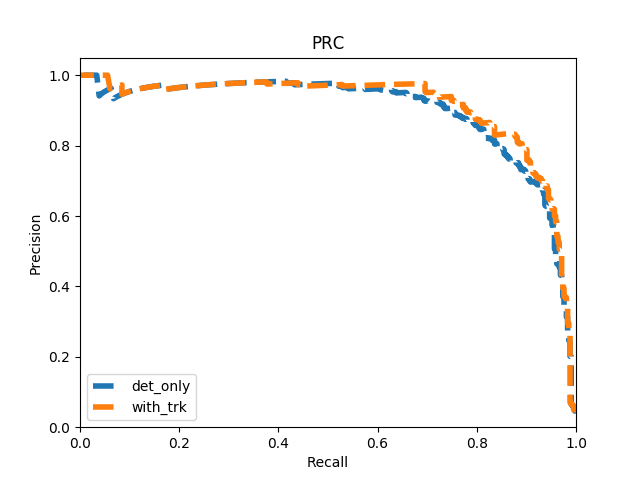}}
    \hspace{0pt}%
    \subfigure[Pedestrians.]{
    \includegraphics[width=.35\textwidth]{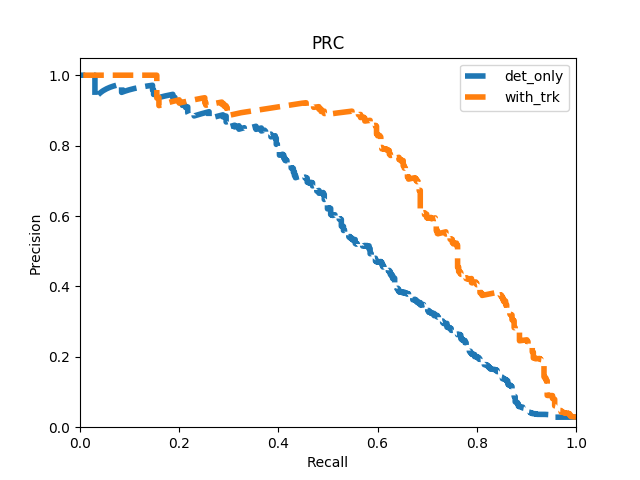}}
\end{adjustwidth}
\caption{PRC for for experiments with real detector and tracker. Notation is the same as Fig. \ref{fig:good_model}.}
\label{fig:real_detector_tracker}
\end{figure*}

\begin{table}[h]
\caption{Performance of the real tracker.}
\label{tab:data_association}
\centering
\begin{adjustwidth}{-0.3 cm}{}
    \begin{tabular}{c|c|c|c|c|c|c|c|c}
    \hline
    Dataset&0001&0005&0014&0019&0020&0035&0059&0084 \\
    \hline
    $N_{object}$& 14 & 12 & 35 & 23 & 7 & 21 & 57 & 53  \\ 
    \hline
    $N_{frame}$& 108 & 154 & 341 & 481 & 86 & 131 & 373 & 383  \\
    \hline
    Mis-association & 0 & 0 & 5 & 3 & 0 & 0 & 23 & 0 \\ \hline
    Discontinuity & 33 & 38 & 26 & 53 & 39 & 47 & 94 & 125  \\ \hline
    $N_{go}$& 5.0 & 4.6 & 7.9 & 7.9 & 5.4 & 5.3 & 9.7 & 8.4  \\
    \hline
    \multicolumn{9}{p{10 cm}}{$N_{object}$: the number of objects; $N_{frame}$: the number of frames, $Mis-association$: the number of wrong data-association; $Discontinuity$: times that the tracking discontinuity happens. $N_{go}$: average life span of a tracklet.}
    \end{tabular}
\end{adjustwidth}
\end{table}

\subsubsection{Efficiency}
The fusion model requires the detection model to run multiple times to reduce the ambiguity in detection. Hence, the computation cost of the actual detection is higher than that of the ideal detector and tracker. The efficiency is depicted in Fig. \ref{fig:efficiency_fusion_ideal}. Compared with the ideal detector, the changes in efficiency along the range of interest are very similar. The larger the range of interest, the more efficient the method will be. However, the computation cost with the real detector is about 10 times as much as the ideal detector. That is, the confusion model selects on average around 10 keyframes from a stream of point clouds to fuse them for better detection accuracy.

\subsection{Experiments with Real Detector and Tracker}

The tracker based on the EKF and Hungarian data association was used to track all objects within the region of interest. The starting point of the robot was set as the origin of the global coordinate framework, and the dead-reckoning algorithm was used to obtain the global coordinate of the robot. This means that all detected objects are transformed into the global coordinate framework from the Lidar local coordinate framework for data association. Compared with simulation-based experiments, this real-world experiment will validate how the tracker can boost efficiency and accuracy without the perfect function of data association and state estimation.

\begin{figure}
    \centering
    \includegraphics[width=.4\textwidth]{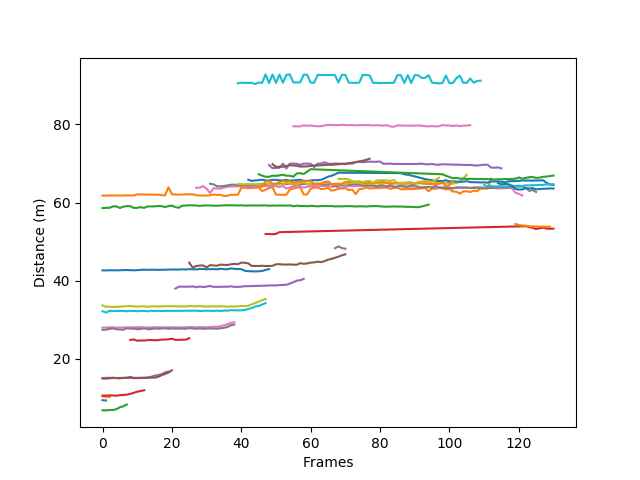}
    \caption{Distance between centre of static proposals and car starting point against the frame ID of dataset 0035.}
    \label{fig:wrong_real_tracker}
\end{figure} 

\subsubsection{Efficiency}
The metric for efficiency improvement is the same as the one used in the simulation. The results can be found in Fig. \ref{fig:efficiency_real_tracker}, which shows that the real tracker can still improve the efficiency greatly, but not as much as in the previous experiments. This is caused by a large number of tracking discontinuities and small $N_{go}$, as shown in Table \ref{tab:data_association}. In our detection method, we average all the points in one proposal to obtain its central coordinate rather than using a complicated method to regress its central location. This could lead to frequent jumps of its central location in the global coordinate framework when the proposal’s views concerning robot changes. Such jumps are apparent in Fig. \ref{fig:wrong_real_tracker}. In that figure, all the objects are static and their distance to the origin in the global coordinate framework across consecutive frames should be flat horizontal lines. However, several lines have random jitters and some are drifting away from the location at the last tracking stage. This can be caused either by our simple method to infer the central location of proposals or if there is an incomplete view of the object (e.g. objects are truncated by the boundary in the field of view). When a proposal moves out of the field of view, this causes a ‘shift effect’ at the end of its tracking curve, see Fig. \ref{fig:wrong_real_tracker}. A jump can fail one data association easily and lead to the introduction of a new tracking stream. This is treated as a tracking discontinuity. For the new tracking stream, the classification model is launched to identify the object as if it were new. The jumpy effect and the number of tracking discontinuities can be reduced by increasing $T_{DA}$, but a large $T_{DA}$ will result in a greater number of mis-associations. If the better tracker is used, the efficiency will be improved and it is closer than our tracker to the results with the ideal tracker.

\subsubsection{Accuracy}
The accuracy performance is reflected in Fig. \ref{fig:real_detector_tracker} and Table \ref{tab:real_detector_tracker}. For the car and the cyclist, the improvement in accuracy from the real tracker is less than that from the ideal tracker (Fig. \ref{fig:good_model}). This could happen if the real tracker did not perform data association well enough, as shown in Table \ref{tab:data_association}. The table shows that the number of tracking discontinuities is large, which could limit the ability of the fusion model to combine the information from different observations of the same object. Moreover, the mis-association of tracking streams, which include objects with different classes, could decrease the detection performance as the fusion model will reduce the class confidence. Despite several tracking discontinuities, some mis-association and a short lifespan in tracking, the real tracker can still make a substantial improvement in the detection performance for the pedestrian. However, the detector fails to detect confidently.

\section{Conclusion}\label{conclusion_part}
Inspired by the pattern that human beings use to perceive their surroundings, we aimed to investigate how a tracker can be applied to improve the efficiency and accuracy of a detector. The corresponding detection framework with tracking was proposed to realize the idea and a substantial number of experiments were carried out to analyse the proposed framework. The results showed that the proposed method outperforms original tracking-by-detection approaches both in, efficiency and accuracy. Even though we believe that this approach can be generalized to all robots, the framework is designed mainly for small robots with limited processing resources. Thus, it cannot be generalized to complicated mobile robotic systems. Therefore, in future work, we will extend this concept to robots with sufficient processing capabilities and design the framework with a detector and tracker based on deep neural networks.

\section{Acknowledgment}
The presented research was supported by the China Scholarship Council (Grant no. 201606950019).

\bibliographystyle{IEEEtran}
\bibliography{IEEEabrv,main}

\begin{IEEEbiography}[{\includegraphics[width=1in,height=1.25in,keepaspectratio]{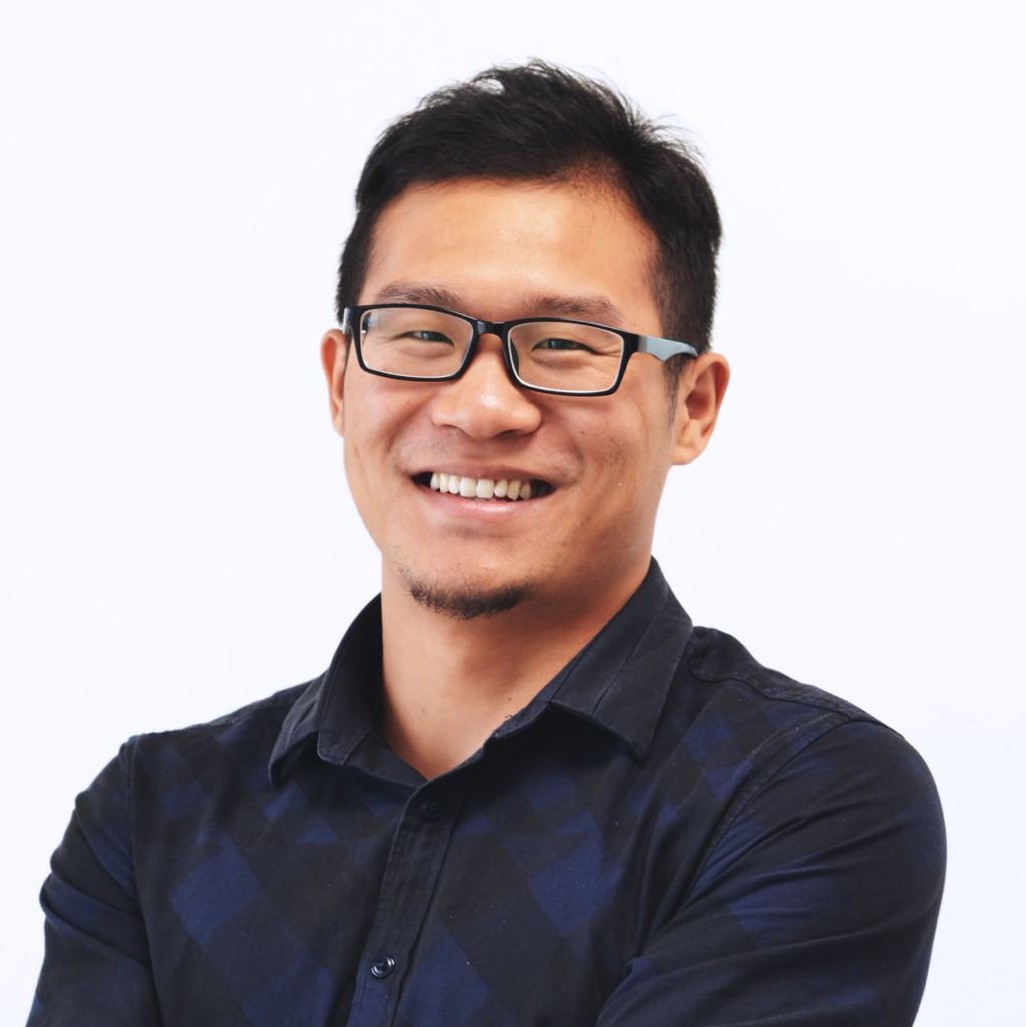}}]{Xuesong LI}
received his M.Sc. degree in Automotive Engineering from Wuhan University of Technology, Hubei, China, in 2016. He is currently a Ph.D. student in Robotics, University of New South Wales, Sydney. His research topic is about perception system for autonomous driving vehicles.
\end{IEEEbiography}

\vskip  -0.001\baselineskip plus -1fil
\begin{IEEEbiography}[{\includegraphics[width=1.in,height=1.25
in,clip,keepaspectratio]{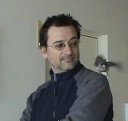}}]{Jose E Guivant}
obtained his Ph.D. degree, in Robotics, from The University of Sydney, Australia, in July/2002. He is currently Sr. Lecturer in Mechatronics, at the School of Mechanical Engineering, University of New South Wales, Australia.
\end{IEEEbiography}

\end{document}